\pdfoutput=1
\documentclass[11pt]{article}

\usepackage[]{acl}

\usepackage{times}
\usepackage{latexsym} 

\usepackage[T1]{fontenc}
\usepackage[utf8]{inputenc}

\usepackage{microtype}

\usepackage{bm}
\usepackage{amsmath}
\usepackage{amssymb}
\usepackage{graphicx}
\usepackage{booktabs}
\usepackage{multirow}

\title{One Reference Is Not Enough: Diverse Distillation with Reference Selection for Non-Autoregressive Translation}

\author{Chenze Shao$^{1,2}$, Xuanfu Wu$^{1,2}$, Yang Feng$^{1,2}$\thanks{\ \ Corresponding author: Yang Feng}\\
$^{1}$ Key Laboratory of Intelligent Information Processing\\
Institute of Computing Technology, Chinese Academy of Sciences (ICT/CAS)\\
$^{2}$ University of Chinese Academy of Sciences\\
{\tt \{\href{mailto:shaochenze18z@ict.ac.cn}{shaochenze18z}, \href{mailto:wuxuanfu20s@ict.ac.cn}{wuxuanfu20s}, \href{mailto:fengyang@ict.ac.cn}{fengyang}\}@ict.ac.cn}}

\begin{document}
\maketitle
\begin{abstract}
Non-autoregressive neural machine translation (NAT) suffers from the multi-modality problem: the source sentence may have multiple correct translations, but the loss function is calculated only according to the reference sentence. Sequence-level knowledge distillation makes the target more deterministic by replacing the target with the output from an autoregressive model. However, the multi-modality problem in the distilled dataset is still nonnegligible. Furthermore, learning from a specific teacher limits the upper bound of the model capability, restricting the potential of NAT models. In this paper, we argue that one reference is not enough and propose diverse distillation with reference selection (DDRS) for NAT. Specifically, we first propose a method called SeedDiv for diverse machine translation, which enables us to generate a dataset containing multiple high-quality reference translations for each source sentence. During the training, we compare the NAT output with all references and select the one that best fits the NAT output to train the model. Experiments on widely-used machine translation benchmarks demonstrate the effectiveness of DDRS, which achieves 29.82 BLEU with only one decoding pass on WMT14 En-De, improving the state-of-the-art performance for NAT by over 1 BLEU.\footnote{Source code: https://github.com/ictnlp/DDRS-NAT.}

\end{abstract}

\section{Introduction}
Non-autoregressive machine translation \cite{gu2017non} has received increasing attention in the field of neural machine translation for the property of parallel decoding. Despite the significant speedup, NAT suffers from the performance degradation compared to autoregressive models \cite{bahdanau2014neural,vaswani2017attention} due to the multi-modality problem: the source sentence may have multiple correct translations, but the loss is calculated only according to the reference sentence. The multi-modality problem will cause the inaccuracy of the loss function since NAT has no prior knowledge about the reference sentence during the generation, where the teacher forcing algorithm \cite{williams1989learning} makes autoregressive models less affected by feeding the golden context.

How to overcome the multi-modality problem has been a central focus in recent efforts for improving NAT models \cite{shao-etal-2019-retrieving,DBLP:conf/aaai/ShaoZFMZ20,DBLP:journals/corr/abs-2106-08122,ran-etal-2020-learning,pmlr-v119-sun20c,Aligned,Du2021OAXE}. A standard approach is to use sequence-level knowledge distillation \cite{kim-rush-2016-sequence}, which attacks the multi-modality problem by replacing the target-side of the training set with the output from an autoregressive model. The distilled dataset is less complex and more deterministic \cite{Zhou2020Understanding}, which becomes a default configuration of NAT. However, the multi-modality problem in the distilled dataset is still nonnegligible \cite{Zhou2020Understanding}. Furthermore, the distillation requires NAT models to imitate the behavior of a specific autoregressive teacher, which limits the upper bound of the model capability and restricts the potential of developing stronger NAT models.

\begin{figure*}[t]
  \begin{center}
    \includegraphics[width=2\columnwidth]{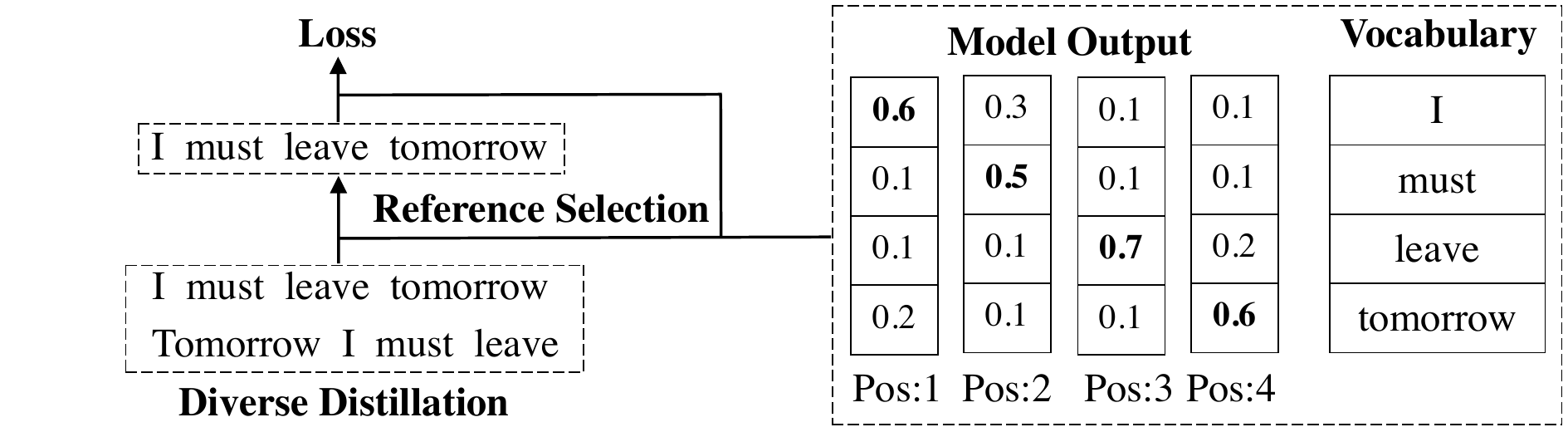}
    \caption{Illustration of diverse distillation and reference selection. Diverse distillation provides multiple references and reference selection selects the one that best fits the model output for the training.}
    \label{fig:1}
  \end{center}
\end{figure*}

In this paper, we argue that one reference is not enough and propose diverse distillation with reference selection (DDRS) for NAT. Diverse distillation generates a dataset containing multiple reference translations for each source sentence, and reference selection finds the reference translation that best fits the model output for the training. 
As illustrated in Figure \ref{fig:1}, diverse distillation provides candidate references ``I must leave tomorrow" and ``Tomorrow I must leave", and reference selection selects the former which fits better with the model output. More importantly, NAT with DDRS does not imitate the behavior of a specific teacher but learns selectively from multiple references, which improves the upper bound of the model capability and allows for developing stronger NAT models.

The object of diverse distillation is similar to the task of diverse machine translation, which aims to generate diverse translations with high translation quality \cite{li2016simple,DBLP:conf/aaai/VijayakumarCSSL18,pmlr-v97-shen19c,wu-etal-2020-generating,li-etal-2021-mixup-decoding}. We propose a simple yet effective method called SeedDiv, which directly uses the randomness in model training controlled by random seeds to produce diverse reference translations without losing translation quality. 
For reference selection, we compare the model output with all references and select the one that best fits the model output, which can be efficiently conducted without extra neural computations. The model learns from all references indiscriminately in the beginning, and gradually focuses more on the selected reference that provides accurate training signals for the model. We also extend the reference selection approach to reinforcement learning, where we encourage the model to move towards the selected reference that gives the maximum reward to the model output. 

We conduct experiments on widely-used machine translation benchmarks to demonstrate the effectiveness of our method. On the competitive task WMT14 En-De, DDRS achieves 27.60 BLEU with $14.7\times$ speedup and 28.33 BLEU with $5.0\times$ speedup, outperforming the autoregressive Transformer while maintaining considerable speedup. When using the larger version of Transformer, DDRS even achieves 29.82 BLEU with only one decoding pass, improving the state-of-the-art performance level for NAT by over 1 BLEU. 

\section{Background}
\subsection{Non-Autoregressive Translation}
\citet{gu2017non} proposes non-autoregressive machine translation to reduce the translation latency through parallel decoding. The vanilla-NAT models the translation probability from the source sentence $\bm{x}$ to the target sentence $\bm{y}\!=\!\{y_1, ..., y{_T}\}$ as:

\begin{equation}
\label{eq:nonauto_prob}
p(\bm{y}|\bm{x},\theta) = \prod_{t=1}^{T}p_t(y_t|\bm{x},\theta),
\end{equation}
where $\theta$ is a set of model parameters and $p_t(y_t|\bm{x},\theta)$ is the translation probability of word $y_t$ in position $t$. The vanilla-NAT is trained to minimize the cross-entropy loss:
\begin{equation}
\begin{aligned}
\label{eq:nonauto_mle}
\mathcal{L}_{CE}(\theta) = -\sum_{t=1}^{T}\log(p_t(y_t|\bm{x},\theta)).
\end{aligned}
\end{equation}

The vanilla-NAT has to know the target length before constructing the decoder inputs. The target length $T$ is set as the reference length during the training and obtained from a length predictor during the inference. The target length cannot be changed dynamically during the inference, so it often requires generating multiple candidates with different lengths and re-scoring them to produce the final translation \cite{gu2017non}. 

The length issue can be overcome by connectionist temporal classification \citep[CTC,][]{10.1145/1143844.1143891}. CTC-based models usually generate a long alignment containing repetitions and blank tokens. The alignment will be post-processed by a collapsing function $\Gamma^{-1}$ to recover a normal sentence, which first collapses consecutive repeated tokens and then removes all blank tokens. CTC is capable of efficiently finding all alignments $\bm{a}$ which the reference sentence $\bm{y}$ can be recovered from, and marginalizing the log-likelihood with dynamic programming:
\begin{equation}
    \log p(\bm{y}|\bm{x},\theta) = \log \sum_{\bm{a} \in \Gamma(\bm{y})} p(\bm{a}|\bm{x},\theta).
\end{equation}
Due to the superior performance and the flexibility of generating predictions with variable length, CTC is receiving increasing attention in non-autoregressive translation \cite{libovicky2018end,kasner2020improving,saharia-etal-2020-non,gu2020fully,zheng2021duplex}.

\subsection{Sequence-Level Knowledge Distillation}
Sequence-level Knowledge Distillation \citep[SeqKD,][]{kim-rush-2016-sequence} is a widely used knowledge distillation method in NMT, which trains the student model to mimic teacher's actions at sequence-level. Given the student prediction $p$ and the teacher prediction $q$, the distillation loss is:
\begin{equation}
\begin{aligned}
\label{eq:skd}
\mathcal{L}_{SeqKD}(\theta)&=-\sum_{\bm{y}}  q(\bm{y}|\bm{x}) \log p(\bm{y}|\bm{x},\theta)\\
&\approx-\log p(\bm{\hat{y}}|\bm{x},\theta),
\end{aligned}
\end{equation}
where $\theta$ are parameters of the student model and $\bm{\hat{y}}$ is the output from running beam search with the teacher model. The teacher output $\bm{\hat{y}}$ is used to approximate the teacher distribution otherwise the distillation loss will be intractable. 

The procedure of sequence-level knowledge distillation is: (1) train a teacher model, (2) run beam search over the training set with this model, (3) train the student model with cross-entropy on the source sentence and teacher translation pairs. The distilled dataset is less complex and more deterministic \cite{Zhou2020Understanding}, which helps to alleviate the multi-modality problem and becomes a default configuration in NAT models.
\subsection{Diverse Machine Translation}
The task of diverse machine translation requires to generate diverse translations and meanwhile maintain high translation quality. Assume the reference sentence is $\bm{y}$ and we have multiple translations $\{\bm{y_1}, ..., \bm{y_k}\}$, the translation quality is measured by the average reference BLEU (rfb):
\begin{equation}
\text{rfb} = \frac{1}{k}\sum_{i=1}^{k}\text{BLEU}(\bm{y},\bm{y_i}),
\end{equation}
and the translation diversity is measured by the average pairwise BLEU (pwb):
\begin{equation}
\text{pwb} = \frac{1}{(k-1)k}\sum_{i=1}^{k}\sum_{j\neq i}\text{BLEU}(\bm{y_i},\bm{y_j}).
\end{equation}

Higher reference BLEU indicates better translation quality and lower pairwise BLEU indicates better translation diversity. Generally speaking, there is a trade-off between quality and diversity. In existing methods, translation diversity has to be achieved at the cost of losing translation quality.
\section{Approach}
In this section, we first introduce the diverse distillation technique we use to generate multiple reference translations for each source sentence, and then apply reference selection to select the reference that best fits the model output for the training. 
\subsection{Diverse Distillation}
The objective of diverse distillation is to obtain a dataset containing multiple high-quality references for each source sentence, which is similar to the task of diverse machine translation that aims to generate diverse translations with high translation quality. However, the translation diversity is achieved at a certain cost of translation quality in previous work, which is not desired in diverse distillation.

Using the randomness in model training, we propose a simple yet effective method called SeedDiv to achieve translation diversity without losing translation quality. Specifically, given the desired number of translations $k$, we directly set $k$ different random seeds to train $k$ translation models, where random seeds control the random factors during the model training such as parameter initialization, batch order, and dropout. During the decoding, each model translates the source sentence with beam search, which gives $k$ different translations in total. Notably, SeedDiv does not sacrifice the translation quality to achieve diversity since random seeds do not affect the expected model performance.

\begin{figure}[t]
  \begin{center}
    \includegraphics[width=0.9\columnwidth]{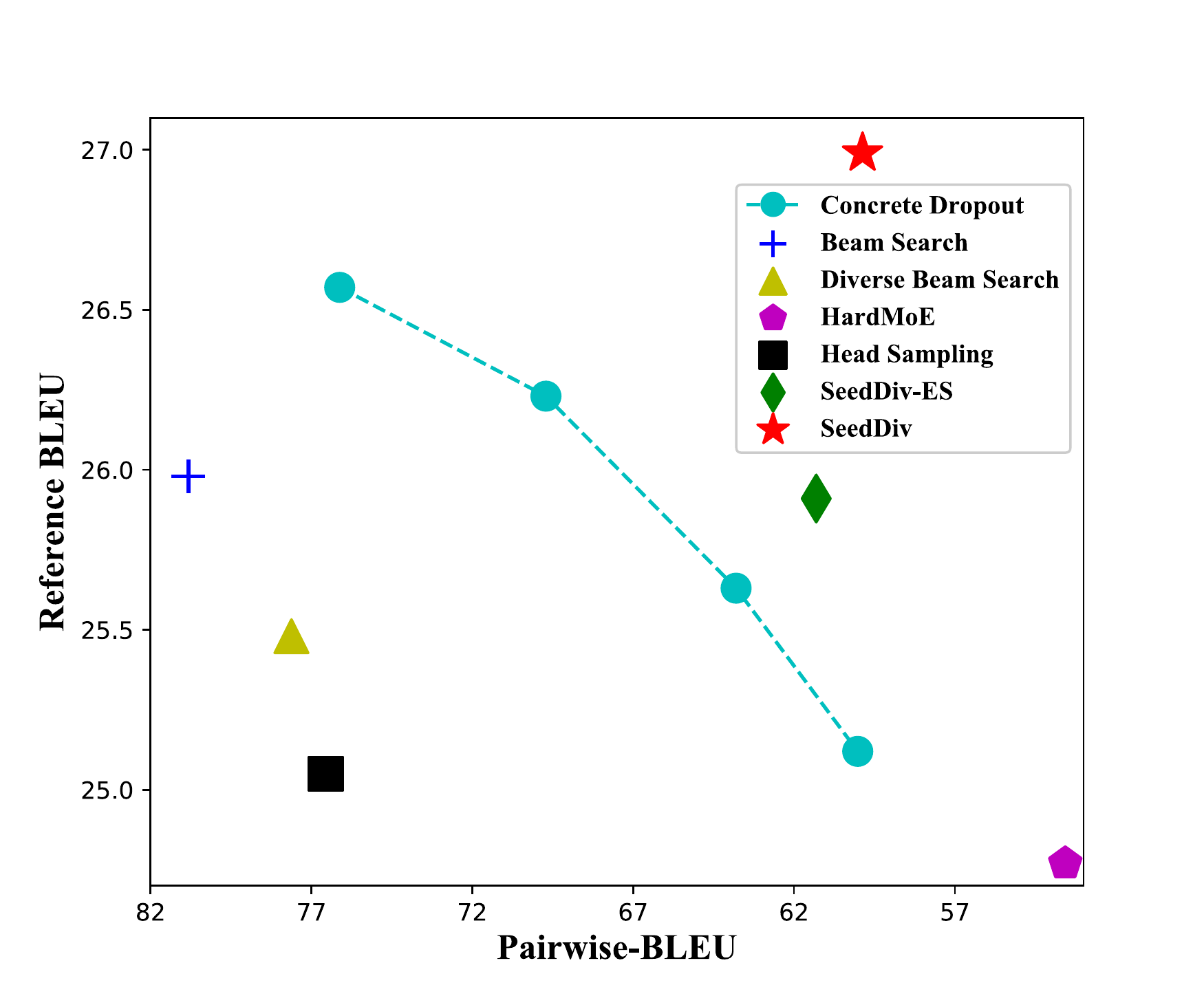}
    \caption{Reference BLEU and pairwise BLEU scores of SeedDiv and other diverse translation methods on the test set of WMT14 En-De. We do not use compound split to keep consistency with previous work.}
    \label{fig:div}
  \end{center}
\end{figure}

We conduct the experiment on WMT14 En-De to evaluate the performance of SeedDiv. We use the base setting of Transformer and train the model for 150K steps. The detailed configuration is described in section \ref{setting}. We also re-implement several existing methods with the same setting for comparison, including Beam Search, Diverse Beam Search \cite{DBLP:conf/aaai/VijayakumarCSSL18}, HardMoE \cite{pmlr-v97-shen19c}, Head Sampling \cite{DBLP:conf/aaai/SunHWDC20} and Concrete Dropout \cite{wu-etal-2020-generating}. We set the number of translations $k\!=\!3$, and set the number of heads to be sampled as 3 for head sampling. We also implement a weaker version of our method SeedDiv-ES, which early stops the training process with only $\frac{1}{k}$ of total training steps. We report the results of these methods in Figure \ref{fig:div}. 

It is surprising to see that SeedDiv achieves outstanding translation diversity besides the superior translation quality, which outperforms most methods on both translation quality and diversity. Only HardMoe has a better pairwise BLEU than SeedDiv, but its reference BLEU is much lower. The only concern is that SeedDiv requires a larger training cost to train multiple models, so we also use a weaker version SeedDiv-ES for comparison. Though the model performance is degraded due to the early stop, SeedDiv-ES still achieves a good trade-off between the translation quality and diversity, demonstrating the advantage of using the training randomness controlled random seeds to generate diverse translations. Therefore, we use SeedDiv as the technique for diverse distillation.

\subsection{Reference Selection}
\subsubsection{Losses under Diverse Distillation}
\label{sec:losses}
After diverse distillation, we obtain a dataset containing $k$ reference sentences $\bm{y_{1:k}}$ for each source sentence $\bm{x}$. Traditional data augmentation algorithms for NMT \cite{sennrich2016improving,zhang2016exploiting,zhou-keung-2020-improving,nguyen2019data} generally calculate cross-entropy losses on all data and use their summation to train the model:
\begin{equation}
\label{eq:sum}
   \mathcal{L}_{sum}(\theta)= -\frac{1}{k}\sum_{i=1}^{k}\log p(\bm{y_i}|\bm{x},\theta).
\end{equation}
However, this loss function is inaccurate for NAT due to the increase of data complexity. Sequence-level knowledge distillation works well on NAT by reducing the complexity of target data \cite{Zhou2020Understanding}. In comparison, the target data generated by diverse distillation is relatively more complex. If NAT learns from the $k$ references indiscriminately, it will not eventually converge to any one reference but generate a mixture of all references. 

Using the multi-reference dataset, we propose to train NAT with reference selection to evaluate the model output with better accuracy. We compare the model output with all reference sentences and select the one with the maximum probability assigned by the model. We train the model with only the selected reference:
\begin{equation}
\label{eq:max}
\mathcal{L}_{max}(\theta)= -\log\max_{1\leq i \leq k} p(\bm{y_i}|\bm{x},\theta).
\end{equation}
In this way, we do not fit the model to all references but only encourage it to generate the nearest reference, which is an easier but more suitable objective for the model. Besides, when the ability of autoregressive teacher is limited, the NAT model can learn to ignore bad references in the data and select the clean reference for the training, which makes the capability of NAT not limited by a specific autoregressive teacher.

In addition to minimizing all losses $\mathcal{L}_{sum}(\theta)$ or the selected loss $\mathcal{L}_{max}(\theta)$, there is also an intermediate choice to assign different weights to reference sentences. We can optimize the log-likelihood of generating any reference sentence as follows:
\begin{equation}
\label{eq:probsum}
\mathcal{L}_{mid}(\theta)= -\log\sum_{i=1}^{k} p(\bm{y_i}|\bm{x},\theta).
\end{equation}
The gradient of Equation \ref{eq:probsum} is equivalent to assigning weight $\frac{p(\bm{y_i}|\bm{x},\theta)}{\sum_{i}p(\bm{y_i}|\bm{x},\theta)}$ to the cross-entropy loss of each reference sentence $\bm{y_i}$. In this way, the model focuses more on suitable references but also assigns non-zero weights to other references.

We use a linear annealing schedule with two stages to train the NAT model. 
In the first stage, we begin with the summation $\mathcal{L}_{sum}(\theta)$ and linearly anneal the loss to $\mathcal{L}_{mid}(\theta)$. Similarly, we linearly switch to the selected loss $\mathcal{L}_{max}(\theta)$ in the second stage.
We use $t$ and $T$ to denote the current time step and total training steps respectively, and use a constant $\lambda$ to denote the length of the first stage. The loss function is:
\begin{equation}
\label{eq:anneal}
\mathcal{L}(\theta)\!=\!\left\{
             \begin{array}{ll}
             \!T_1\mathcal{L}_{mid}(\theta)\!+\!(1\!-\!T_1)\mathcal{L}_{sum}(\theta),\!&\!t\!\leq\!\lambda T \\
             \!T_2\mathcal{L}_{max}(\theta)\!+\!(1\!-\!T_2)\mathcal{L}_{mid}(\theta),\!&\!t\!>\!\lambda T \\
             \end{array}
\!,\right.
\end{equation}
where $T_1$ and $T_2$ are defined as:
\begin{equation}
T_1 = \frac{t}{\lambda T},\quad T_2=\frac{t-\lambda T}{T-\lambda T}. 
\end{equation}
In this way, the model learns from all references indiscriminately at the beginning, which serves as a pretraining stage that provides comprehensive knowledge to the model. As the training progresses, the model focuses more on the selected reference, which provides accurate training signals and gradually finetunes the model to the optimal state.
\subsubsection{Efficient Calculation with CTC}
To calculate the probability $p(\bm{y}|\bm{x},\theta)$, the vanilla-NAT must set the decoder length to the length of $\bm{y}$. Therefore, calculating the probability of $k$ reference sentences requires running the decoder for at most $k$ times, which will greatly increase the training cost. Fortunately, for CTC-based NAT, the training cost is nearly the same since its decoder length is only determined by the source sentence. We only need to run the model once and calculate the probabilities of the $k$ reference sentences with dynamic programming, which has a minor cost compared with forward and backward propagations. In Table \ref{tab:forback}, we show the calculation cost of $\mathcal{L}_{max}(\theta)$ and $\mathcal{L}_{sum}(\theta)$ for different models. We use CTC as the baseline model due to its superior performance and training efficiency.

\begin{table}[t]
\centering
\begin{tabular}{lcccc}
\toprule
 \multirow{2}{*}{Models}&\multicolumn{2}{c}{$\mathcal{L}_{max}(\theta)$} & \multicolumn{2}{c}{$\mathcal{L}_{sum}(\theta)$} \\
 &For & Back & For & Back\\
\midrule
AT & $k\times$ & $1\times$ & $k\times$ & $k\times$\\
vanilla-NAT & $k\times$ &$1\times$  & $k\times$ & $k\times$ \\
CTC & $1\times$ & $1\times$ & $1\times$ & $1\times$ \\
\bottomrule
\end{tabular}
\caption{The calculation cost of $\mathcal{L}_{max}(\theta)$ and $\mathcal{L}_{sum}(\theta)$ for different models. `For' and `Back' indicate forward and backward propagations respectively.}
\label{tab:forback}
\end{table}

\subsubsection{Max-Reward Reinforcement Learning}
Following \citet{shao-etal-2019-retrieving,DBLP:journals/corr/abs-2106-08122}, we finetune the NAT model with the reinforcement learning objective \cite{williams1992simple,ranzato2015sequence}:
\begin{equation}
\label{eq:rl-1}
\mathcal{L}_{rl}(\theta)=\mathop{\mathbb{E}}_{\bm{y}}[\log p(\bm{y}|\bm{x},\theta) \cdot r(\bm{y})],
\end{equation}
where $r(\bm{y})$ is the reward function and will be discussed later. The usual practice is to sample a sentence $\bm{y}$ from the distribution $p(\bm{y}|\bm{x},\theta)$ to estimate the above equation. For CTC based NAT, $p(\bm{y}|\bm{x},\theta)$ cannot be directly sampled, so we sample from the equivalent distribution $p(\bm{a}|\bm{x},\theta)$ instead. We recover the target sentence by the collapsing function $\Gamma^{-1}$ and calculate its probability with dynamic programming to estimate the following equation:
\begin{equation}
\label{eq:rl-2}
\mathcal{L}_{rl}(\theta)=\mathop{\mathbb{E}}_{\bm{a}}[\log p(\Gamma^{-1}(\bm{a})|\bm{x},\theta) \cdot r(\Gamma^{-1}(\bm{a}))].
\end{equation}

The reward function is usually evaluation metrics for machine translation (e.g., BLEU, GLEU), which evaluate the prediction by comparing it with the reference sentence. We use $r(\bm{y_1},\bm{y_2})$ to denote the reward of prediction $\bm{y_1}$ when $\bm{y_2}$ is the reference. As we have $k$ references $\bm{y_{1:k}}$, we define our reward function to be the maximum reward:
\begin{equation}
\label{eq:reward}
r(\bm{y})=\max_{1\leq i\leq k}r(\bm{y},\bm{y_i}).
\end{equation}
By optimizing the maximum reward, we encourage the model to move towards the selected reference, which is the closest to the model. Otherwise, rewards provided by other references may mislead the model to generate a mixture of all references.
\section{Experiments}
\subsection{Experimental Settings}
\label{setting}
\begin{table*}[th!]
\centering
\small
\begin{tabular}{clcccccc}
\toprule
 \multirow{2}{*}{{Models}} &&
 \multirow{2}{*}{{Iterations}} &
 \multirow{2}{*}{{Speedup}}&
 \multicolumn{2}{c}{{WMT14}} & \multicolumn{2}{c}{{WMT16}} \\
 \multicolumn{2}{c}{} & & & {EN-DE} & {DE-EN} & {EN-RO} & {RO-EN} \\
\midrule
\multirow{2}{*}{AT}
& Transformer \cite{vaswani2017attention}& N & 1.0$\times$ &  27.51&31.52&34.39&33.76 \\
& \quad + distillation ($k$=3)& N & 1.0$\times$ &  \bf{28.04}&\bf{32.17}&\bf{35.10}&\bf{34.83} \\
\cmidrule[0.6pt](lr){1-8}
&NAT-FT \cite{gu2017non} &1& 15.6$\times$ &17.69 & 21.47 & 27.29 &  29.06\\
& CTC~\cite{libovicky2018end} & 1 & -- & 16.56 & 18.64 & 19.54 & 24.67 \\
& NAT-REG \cite{wang2019non} & 1 & 27.6$\times$ & 20.65 & 24.77 & -- & --\\
&Bag-of-ngrams \cite{DBLP:conf/aaai/ShaoZFMZ20} &1&10.7$\times$&20.90 &24.61 &28.31 &29.29\\
& AXE \cite{Aligned}& 1 & -- & 23.53 & 27.90 & 30.75 & 31.54 \\
&SNAT \cite{liu-etal-2021-enriching} & 1 & 22.6$\times$ & 24.64 & 28.42 & 32.87 & 32.21 \\
&GLAT \cite{qian-etal-2021-glancing} & 1 & 15.3$\times$&25.21&29.84&31.19&32.04\\
One-pass&Seq-NAT \cite{DBLP:journals/corr/abs-2106-08122} & 1 & 15.6$\times$&25.54&29.91&31.69&31.78\\
NAT&CNAT \cite{bao-etal-2021-non} & 1 & 10.37$\times$&25.56&29.36&--&--\\
& Imputer \cite{saharia-etal-2020-non}& 1 & --  & 25.80 & 28.40 & 32.30 & 31.70 \\
&OAXE \cite{Du2021OAXE} & 1&--&26.10 & 30.20 & 32.40 & 33.30\\
&AligNART \cite{song-etal-2021-alignart} & 1 & 13.4$\times$&26.40&30.40&32.50&33.10\\
&REDER \cite{zheng2021duplex} & 1 & 15.5$\times$ &26.70 & 30.68 & 33.10 & 33.23\\
&CTC w/ DSLP\&MT \cite{Huang2021NonAutoregressiveTW} & 1 & 14.8$\times$&27.02&31.61&34.17&34.60\\
& Fully-NAT \cite{gu2020fully} & 1 & 16.8$\times$ & 27.20 & \bf{31.39} & \bf{33.71} & \bf{34.16}\\
&REDER + beam20 + AT reranking& 1 & 5.5$\times$ &\bf{27.36} & 31.10 & 33.60 & 34.03\\
\cmidrule[0.6pt](lr){1-8}
& iNAT \cite{lee2018deterministic} & 10 & 2.0$\times$ & 21.61 & 25.48 & 29.32 & 30.19 \\
& CMLM \cite{ghazvininejad2019maskpredict} & 10 & -- & 27.03 & 30.53 & 33.08 & 33.31 \\
& RecoverSAT \cite{ran-etal-2020-learning} & N/2 & $2.1\times$ & 27.11 & 31.67 & 32.92 & 33.19 \\
Iterative& LevT \cite{gu2019levenshtein} & 2.05 & 4.0$\times$ & 27.27 & -- & -- & 33.26 \\
NAT& DisCO \cite{Kasai2020NonautoregressiveMT} & 4.82 & -- & 27.34 & 31.31 & 33.22 & 33.25  \\
&JM-NAT \cite{guo-etal-2020-jointly} & 10 & -- & 27.69 & \bf{32.24} & 33.52 & 33.72 \\
&RewriteNAT \cite{geng-etal-2021-learning} & 2.70 & -- & 27.83 & 31.52 & 33.63 & 34.09 \\
& Imputer \cite{saharia-etal-2020-non} & 8& -- & \bf{28.20} & 31.80 & \bf{34.40} & \bf{34.10} \\
\cmidrule[0.6pt](lr){1-8}
\multirow{5}{*}{Our work}
& CTC & 1 & 14.7$\times$& 26.09 & 29.50 & 33.55 & 32.98\\
& CTC + distillation ($k$=3)& 1 & 14.7$\times$& 26.35 & 29.73 & 33.51 & 32.82\\
& DDRS w/o RL& 1 & 14.7$\times$& 27.18 & 30.91 & 34.42 & 34.31\\
& DDRS & 1 & 14.7$\times$& 27.60 & 31.48 & 34.60 & 34.65\\
& DDRS + beam20 + 4-gram LM& 1 & 5.0$\times$& \bf{28.33} & \bf{32.43} & \bf{35.42} & \bf{35.81}\\

\bottomrule
\end{tabular}
\caption{Performance comparison between our models and existing methods. The speedup is measured on WMT14 En-De test set. N denotes the length of translation. $k$ means ensemble distillation \cite{freitag2017ensemble} from an ensemble of $k$ AT models. -- means not reported.}
\label{tab:main-rst}
\end{table*}
\noindent{}\textbf{Datasets} We conduct experiments on major benchmark datasets for NAT: WMT14 English$\leftrightarrow$German (En$\leftrightarrow$De, 4.5M sentence pairs) and WMT16 English$\leftrightarrow$Romanian (En$\leftrightarrow$Ro, 0.6M sentence pairs). We also evaluate our approach on a large-scale dataset WMT14 English$\rightarrow$French (En$\rightarrow$Fr, 23.7M sentence pairs) and a small-scale dataset IWSLT14 German$\rightarrow$English (De$\rightarrow$En, 160K sentence pairs).
The datasets are tokenized into subword units using a joint BPE model \cite{sennrich-etal-2016-neural}.  We use BLEU \cite{papineni2002bleu} to evaluate the translation quality. 

\vspace{5pt}
\noindent{}\textbf{Hyperparameters} We use 3 teachers for diverse distillation and set the seed to $i$ when training the $i$-th teacher. We set the first stage length $\lambda$ to $2/3$. We use sentence-level BLEU as the reward. We adopt Transformer-base \cite{vaswani2017attention} as our autoregressive baseline as well as the teacher model. The NAT model shares the same architecture as Transformer-base.
We uniformly copy encoder outputs to construct decoder inputs, where the length of decoder inputs is $3\times$ as long as the source length. All models are optimized with Adam \cite{DBLP:journals/corr/KingmaB14} with $\beta=(0.9,0.98)$ and $\epsilon=10^{-8}$, and each batch contains approximately 32K source words. On WMT14 En$\leftrightarrow$De and WMT14 En$\rightarrow$Fr, we train AT for 150K steps and train NAT for 300K steps with dropout 0.2. On WMT16 En$\leftrightarrow$Ro and IWSLT14 De$\rightarrow$En, we train AT for 18K steps and train NAT for 150K steps with dropout 0.3. We finetune NAT for 3K steps. The learning rate warms up to $5\cdot10^{-4}$ within 10K steps in pretraining and warms up to $2\cdot10^{-5}$ within 500 steps in RL fine-tuning, and then decays with the inverse square-root schedule. We average the last 5 checkpoints to obtain the final model.
We use GeForce RTX $3090$ GPU for the training and inference. We implement our models based on fairseq \cite{ott2019fairseq}.

\vspace{5pt}
\noindent{}\textbf{Knowledge Distillation} For baseline NAT models, we follow previous works on NAT to apply sequence-level knowledge distillation \cite{kim-rush-2016-sequence} to make the target more deterministic. Our method applies diverse distillation with $k=3$ by default, that is, we use SeedDiv to generate 3 reference sentences for each source sentence.

\vspace{5pt}
\noindent{}\textbf{Beam Search Decoding} For autoregressive models, we use beam search with beam width 5 for the inference. For NAT, the most straightforward way is to generate the sequence with the highest probability at each position. Furthermore, CTC-based models also support beam search decoding optionally combined with n-gram language models \cite{kasner2020improving}. Following \citet{gu2020fully}, we use beam width 20 combined with a 4-gram language model to search the target sentence, which can be implemented efficiently in C++\footnote{https://github.com/parlance/ctcdecode}.

\subsection{Main Results}
We compare the performance of DDRS and existing methods in Table \ref{tab:main-rst}. Compared with the competitive CTC baseline, DDRS achieves a strong improvement of more than 1.5 BLEU on average, demonstrating the effectiveness of diverse distillation and reference selection. Compared with existing methods, DDRS beats the state-of-the-art for one-pass NAT on all benchmarks and beats the autoregressive Transformer on most benchmarks with 14.7$\times$ speedup over it. The performance of DDRS is further boosted by beam search and 4-gram language model, which even outperforms all iterative NAT models with only one-pass decoding. Notably, on WMT16 En$\leftrightarrow$Ro, our method improves state-of-the-art performance levels for NAT by over 1 BLEU. Compared with autoregressive models, our method outperforms the Transformer with knowledge distillation, and meanwhile maintains 5.0$\times$ speedup over it. 

We further explore the capability of DDRS with a larger model size and stronger teacher models. We use the big version of Transformer for distillation, and also add 3 right-to-left (R2L) teachers to enrich the references. We respectively use Transformer-base and Transformer-big as the NAT architecture and report the performance of DDRS in Table \ref{tab:big}. Surprisingly, the performance of DDRS can be further greatly boosted by using a larger model size and stronger teachers. DDRS-big with beam search achieves 29.82 BLEU on WMT14 En-De, which is close to the state-of-the-art performance of autoregressive models on this competitive dataset and improves the state-of-the-art performance for NAT by over 1 BLEU with only one-pass decoding.

\begin{table}[t]
\small
\centering
\begin{tabular}{lcc}
\toprule
{{Models}} & {{BLEU}} &{{Speedup}}\\ 
\midrule
Transformer-base & 27.51 & 1.0$\times$ \\
Transformer-big (teacher) & \bf{28.64} & 0.9$\times$ \\
R2L-Transformer-big (teacher) & 27.96 & 0.9$\times$ \\
\midrule
DDRS-base & 27.99 & 14.7$\times$  \\
\ \ +beam20\&lm & 28.95 & 5.0$\times$  \\
DDRS-big & 28.84 & 14.1$\times$  \\
\ \ +beam20\&lm & \bf{29.82} &  4.8$\times$ \\
\bottomrule
\end{tabular}
\caption{Performance of DDRS on the test set of WMT14 En-De with Transformer-big for distillation.}
\label{tab:big}
\end{table}
\begin{table}[t]
\small
\centering
\begin{tabular}{lcc}
\toprule
{{Models}} & {En-Fr} &{De-En}\\ 
\midrule
Transformer & 40.15 & 34.17 \\
\midrule
CTC & 38.40 & 31.37  \\
DDRS & 39.91 & 33.12  \\
DDRS +beam20\&lm & \bf{40.59} &  \bf{34.74} \\
\bottomrule
\end{tabular}
\caption{Performance of DDRS on the test sets of WMT14 En-Fr and IWSLT14 De-En.}
\label{tab:scale}
\end{table}
We also evaluate our approach on a large-scale dataset WMT14 En-Fr and a small-scale dataset IWSLT14 De-En. Table \ref{tab:scale} shows that DDRS still achieves considerable improvements over the CTC baseline and DDRS with beam search can outperform the autoregressive Transformer.
\subsection{Ablation Study}
\begin{table}[t]
    \small
    \centering
    \begin{tabular}{ccc|c|c|cc}
        \toprule
        $\!\!\!\mathcal{L}_{sum}\!\!\!$ & $\!\!\!\mathcal{L}_{mid}\!\!\!$ & $\!\!\!\mathcal{L}_{max}\!\!\!$ & $\!\lambda\!$ & Reward  & BLEU$_1\!\!$& BLEU$_2$\\
        \midrule
         \checkmark & & & & & 24.61 &25.97 \\
        & \checkmark & & & &  25.23 & \bf{26.90}\\
        & &\checkmark & & &  \bf{25.31}&26.88 \\
                \midrule
        &\checkmark &\checkmark &$0$ & &25.41&26.99\\
        \checkmark&\checkmark &\checkmark &$1/3$ && 25.48&27.13\\
        \checkmark&\checkmark &\checkmark &$2/3$ && \bf{25.59}&\bf{27.18}\\
        \checkmark&\checkmark & &$1$ & &25.45&27.09\\
        \midrule
        \checkmark&\checkmark &\checkmark &$2/3$ &\textit{random}& 25.63&27.26\\
        \checkmark&\checkmark &\checkmark &$2/3$ &\textit{average}& 25.79&27.51\\
        \checkmark&\checkmark &\checkmark &$2/3$ &\textit{maximum}& \bf{25.92}&\bf{27.60}\\
    \bottomrule
    \end{tabular}
        \caption{Ablation study on WMT14 En-De with different combinations of techniques. BLEU$_1$ is the BLEU score on validation set. BLEU$_2$ is the BLEU score on test set. The validation performance of CTC baseline is 24.57 BLEU. $\lambda$ is the length of the first training stage. \textit{random} means the reward of a random reference, \textit{average} means the average reward, and \textit{maximum} means the maximum reward among all references.}
    \label{tab:ablation}
\end{table}
In Table \ref{tab:ablation}, we conduct an ablation study to analyze the effect of techniques used in DDRS. First, we separately use the loss functions defined in Equation \ref{eq:sum}, Equation \ref{eq:max} and Equation \ref{eq:probsum} to train the model. The summation loss $\mathcal{L}_{sum}(\theta)$ has a similar performance to the CTC baseline, showing that simply using multiple references is not helpful for NAT due to the increase of data complexity. The other two losses $\mathcal{L}_{mid}(\theta)$ and $\mathcal{L}_{max}(\theta)$ achieve considerable improvements to the CTC baseline, demonstrating the effectiveness of reference selection. 

Then we use different $\lambda$ to verify the effect of the annealing schedule. With the annealing schedule, the loss is a combination of the three losses but performs better than each of them. Though the summation loss $\mathcal{L}_{sum}(\theta)$ does not perform well when used separately, it can play the role of pretraining and improve the final performance. When $\lambda$ is $2/3$, the annealing schedule performs the best and improves $\mathcal{L}_{max}(\theta)$ by about 0.3 BLEU. 

Finally, we verify the effect of the reward function during the fine-tuning. When choosing a random reference to calculate the reward, the fine-tuning barely brings improvement to the model. The average reward is better than the random reward, and the maximum reward provided by the selected reference performs the best.

\subsection{DDRS on Autoregressive Transformer}
Though DDRS is proposed to alleviate the multi-modality problem for NAT, it can also be applied to autoregressive models. In Table \ref{tab:ddrsat}, we report the performance of the autoregressive Transformer when trained by the proposed DDRS losses. In contrast to NAT, AT prefers the summation loss $\mathcal{L}_{sum}$, and the other two losses based on reference selection even degrade the AT performance. 

It is within our expectation that AT models do not benefit much from reference selection. NAT generates the whole sentence simultaneously without any prior knowledge about the reference sentence, so the reference may not fit the NAT output well, in which case DDRS is helpful by selecting an appropriate reference for the training. In comparison, AT models generally apply the teacher forcing algorithm \cite{williams1989learning} for the training, which feeds the golden context to guide the generation of the reference sentence. With teacher forcing, AT models do not suffer much from the multi-modality problem and therefore do not need reference selection. Besides, as shown in Table \ref{tab:forback}, another disadvantage is that the training cost of DDRS is nearly k times as large, so we do not recommend applying DDRS on AT.

\begin{table}[t]
\small
\centering
\begin{tabular}{ccccc}
\toprule
 Models& $\mathcal{L}_{CE}$ & $\mathcal{L}_{sum}$ & $\mathcal{L}_{mid}$ & $\mathcal{L}_{max}$ \\
\midrule
AT&27.70&28.08&27.37&27.21\\
NAT&26.09&25.97&26.90&26.88\\
\bottomrule
\end{tabular}
\caption{The performance of AT and CTC-based NAT on the same diverse distillation dataset of WMT14 En-De with different loss functions. $\mathcal{L}_{CE}$ is the cross-entropy loss with sequence-level distillation. $\mathcal{L}_{sum}$, $\mathcal{L}_{mid}$, and $\mathcal{L}_{max}$ described in section \ref{sec:losses} are losses for the diverse distillation dataset.}
\label{tab:ddrsat}
\end{table}
\begin{table}[t]
\small
\centering
\begin{tabular}{lccc}
\toprule
Methods & pwb $\Downarrow$& rfb$\Uparrow$&{{BLEU}} \\ 
\midrule
HardMoe&53.57&24.77&24.51\\
Dropout&69.71&26.23&25.35\\
SeedDiv&59.87&26.99&\bf{25.92} \\
\bottomrule
\end{tabular}
\caption{Pairwise BLEU (pwb) and reference BLEU (rfb) scores of diverse translation techniques and their DDRS performance on WMT14 En-De validation set. pwb and rfb scores are measured on WMT14 En-De test set without compound split.}
\label{tab:dd}
\end{table}
\subsection{Effect of Diverse Distillation}
In the diverse distillation part of DDRS, we apply SeedDiv to generate multiple references. There are also other diverse translation techniques that can be used for diverse distillation. In this section, we evaluate the effect of diverse distillation techniques on the performance of DDRS. Besides SeedDiv, we also use HardMoe \cite{pmlr-v97-shen19c} and Concrete Dropout \cite{wu-etal-2020-generating} to generate multiple references, and report their performance in Table \ref{tab:dd}. When applying other techniques for diverse distillation, the performance of DDRS significantly decreases. The performance degradation indicates the importance of high reference BLEU in diverse distillation, as the NAT student directly learns from the generated references.

\subsection{Effect of Reward} 
There are many automatic metrics to evaluate the translation quality. To measure the effect of reward, we respectively use different automatic metrics as reward for RL, which include traditional metrics (BLEU \cite{papineni2002bleu}, METEOR \cite{banerjee-lavie-2005-meteor}, GLEU \cite{wu2016google}) and pretraining-based metrics (BERTScore \cite{Zhang2020BERTScore:}, BLEURT \cite{sellam-etal-2020-bleurt}). We report the results in Table \ref{tab:reward}. Comparing the three traditional metrics, we can see that there is no significant difference in their performance. The two pretraining-based metrics only perform slightly better than traditional metrics. Considering the performance and computational cost, we use the traditional metric BLEU as the reward.

\begin{table}[t]
\centering
\small
\begin{tabular}{lcccc}
\toprule
\multirow{2}{*}{Models}
&\multicolumn{2}{c}{{WMT14}} & \multicolumn{2}{c}{{WMT16}} \\
 & {EN-DE} & {DE-EN} & {EN-RO} & {RO-EN} \\
\midrule
CTC w/o RL & 26.09 & 29.50 & 33.55 & 32.98\\
BLEU & 26.48 & 30.02 & 33.64 & 33.31\\
METEOR & 26.44 & 29.95 & 33.68 & 33.25\\
GLEU & 26.58 & 29.96 & 33.59 & 33.34\\
BERTScore & 26.51 & 30.20 & 33.69 & 33.42\\
BLEURT & 26.66 & 30.05 & 33.71 & 33.35\\
\bottomrule
\end{tabular}
\caption{BLEU scores on WMT test sets when using different automatic metrics as reward to finetune CTC.}
\label{tab:reward}
\end{table}
\subsection{Number of References}
In this section, we evaluate how the number of references affects the DDRS performance. We set the number of references $k$ to different values and train the CTC model with reference selection. We report the performance of DDRS with different $k$ in Table \ref{tab:num}. The improvement brought by increasing $k$ is considerable when $k$ is small, but it soon becomes marginal. Therefore, it is reasonable to use a middle number of references like $k\!=\!3$ to balance the distillation cost and performance.

\begin{table}[t]
\small
\centering
\begin{tabular}{c|c|c|c|c|c}
\toprule
$k$ & 1& 2&3& 5& 8 \\ 
\midrule
BLEU&25.0&25.6&25.9&26.0&26.1\\
\bottomrule
\end{tabular}
\caption{Performance of DDRS with diffent number of references $k$ on WMT14 En-De validation set.}
\label{tab:num}
\end{table}
\subsection{Time Cost}
The cost of preparing the training data is larger for DDRS since it requires training $k$ teacher models and using each model to decode the training set. We argue that the cost is acceptable since the distillation cost is minor compared to the training cost of NAT, and we can reduce the training cost to make up for it. In Table \ref{tab:costs}, we report the performance and time cost of models with different batch sizes on the test set of WMT14 En-De. DDRS makes up for the larger distillation cost by using a smaller training batch, which has a similar cost to the CTC model of 64K batch and achieves superior performance compared to models of 128K batch.

\begin{table}[t]
\small
\centering
\begin{tabular}{lcccc}
\toprule
{{Models}} & {{Distill}}& {{Train}}& Total&{{BLEU}} \\ 
\midrule
CTC (64K)& 5.5h &26.4h &31.9h&26.34\\
CTC (128K)&5.5h&52.5h&58.0h&26.59\\
DDRS (32K)& 16.5h & 19.3h&35.8h&\bf{27.60} \\
\bottomrule
\end{tabular}
\caption{Performance and time cost of models with different batch sizes on WMT14 En-De. The time cost is measured on 8 GeForce RTX 3090 GPUs. The cost of Distill includes training teacher models and decoding source sentences.}
\label{tab:costs}
\end{table}
\section{Related Work}
\citet{gu2017non} proposes non-autoregressive translation to reduce the translation latency, which suffers from the multi-modality problem. A line of work introduces latent variables to model the nondeterminism in the translation process, where latent variables are based on fertilities \cite{gu2017non}, vector quantization \cite{kaiser2018fast,roy2018theory,bao-etal-2021-non} and variational inference \cite{Ma_2019,Shu2020LatentVariableNN}. Another branch of work proposes training objectives that are less influenced by the multi-modality problem to train NAT models \cite{wang2019non,shao-etal-2019-retrieving,DBLP:conf/aaai/ShaoZFMZ20,DBLP:journals/corr/abs-2106-08122,NIPS2019_8566,Aligned,shan2021modeling,Du2021OAXE}. Some researchers consider transferring the knowledge from autoregressive models to NAT \cite{li-etal-2019-hint,wei-etal-2019-imitation,fine,Zhou2020Understanding,pmlr-v119-sun20c}. Besides, some work propose iterative NAT models that refine the model outputs with multi-pass iterative decoding \cite{lee2018deterministic,gu2019levenshtein,ghazvininejad2019maskpredict,ran-etal-2020-learning,Kasai2020NonautoregressiveMT}. Our work is most related to CTC-based NAT models \cite{10.1145/1143844.1143891,libovicky2018end,kasner2020improving,saharia-etal-2020-non,zheng2021duplex,gu2020fully}, which apply the CTC loss to model latent alignments for NAT. In autoregressive models, translations different from the reference can be evaluated with reinforcement learning \cite{ranzato2015sequence,DBLP:conf/nips/NorouziBCJSWS16}, probabilistic n-gram matching \cite{shao2018greedy}, or an evaluation module \cite{DBLP:conf/aaai/FengXGSZY020}.

Our work is also related to the task of diverse machine translation. \citet{li2016simple,DBLP:conf/aaai/VijayakumarCSSL18} adjust the beam search algorithm by introducing regularization terms to encourage generating diverse outputs. \citet{he-etal-2018-sequence,pmlr-v97-shen19c} introduce latent variables with the mixture of experts method and use different latent variables to generate diverse translations. \citet{DBLP:conf/aaai/SunHWDC20} generates diverse translations by sampling different attention heads. \citet{wu-etal-2020-generating} train the translation model with concrete dropout and samples different models from a posterior distribution. \citet{li-etal-2021-mixup-decoding} generate different translations for the input sentence by mixing it with different sentence pairs sampled from the training corpus. \citet{nguyen2019data} augment the training set by translating the source-side and target-side data with multiple translation models, but they do not evaluate the diversity of the augmented data.

\section{Conclusion}
In this paper, we propose diverse distillation with reference selection (DDRS) for NAT. Diverse distillation generates a dataset containing multiple references for each source sentence, and reference selection finds the best reference for the training. DDRS demonstrates its effectiveness on various benchmarks, setting new state-of-the-art performance levels for non-autoregressive translation.
\section{Acknowledgement}
We thank the anonymous reviewers for their insightful comments. This work was supported by National Key R\&D Program of China (NO.2017YFE9132900).
\bibliography{acl}
\bibliographystyle{acl_natbib}
\end{document}